\documentclass[]{article}

\usepackage{arxiv}

\usepackage{mathtools}  
\usepackage{amssymb}    
\usepackage{dsfont}     
\usepackage{amsthm}     

\usepackage{hyperref}
\usepackage[T1]{fontenc}
\usepackage{algpseudocode}
\usepackage{algorithm}
\usepackage{booktabs} 
\usepackage{caption}  
\usepackage{graphicx}
\usepackage{color}

\begin{document}

\author{
Liam Wigney\\
Optimisation and Logistics\\
School of Computer Science \\ and Information Technology\\
Adelaide University\\
Adelaide, Australia
\And
Aneta Neumann\\
Optimisation and Logistics\\
School of Computer Science \\ and Information Technology\\
Adelaide University\\
Adelaide, Australia
\AND
Yew-Soon Ong\\
College of Computing \& Data Science \\
Nanyang Technological University\\
Singapore, Singapore
\And
Frank Neumann\\
Optimisation and Logistics\\
School of Computer Science \\ and Information Technology\\
Adelaide University\\
Adelaide, Australia
}

\title{Insights from Multi-tasking the EAX Algorithm for the Travelling Salesperson Problem}

\maketitle              
\begin{abstract}
Evolutionary multitasking allows several related problems to be solved in a single run of an algorithm. In this paper, we investigate integrating evolutionary multitasking with Edge Assembly Crossover (MT-EAX) to solve the classical Travelling Salesperson Problem (TSP). To fairly compare MT-EAX against standard EAX under strict compute budgets, we evaluate three scaling methods: generation scaling, population scaling, and balanced scaling. Our results show that generationally scaled MT-EAX is highly effective compute-wise in the early stages of the search, saving $60\%$ to $90\%$ of compute for equal or better solution quality. We observe that instance geometry has a significant impact, with clustered, normally distributed instances securing larger improvements than uniformly distributed ones. However, when scaling by population or utilising explicit solution transfer, the results are negative due to population starvation and incompatible cross-instance parent selection. We demonstrate that the advantage of MT-EAX derives from increased diversity through parallel search in early generations, which can be successfully preserved using a decoupled configuration to often strictly outperform or match standard EAX performance at final convergence.

\end{abstract}

\section{Introduction}

In recent years, there has been a significant increase in attention towards evolutionary multitasking. This is a framework where several often related problems can be solved in a single run of an algorithm using its population, rather than by solving the single problems individually. It has been shown to work in a broad range of areas, such as continuous and combinatorial optimisation as well as genetic programming. Key literature includes the book by \cite{feng2023evolutionary} and the special issue edited by \cite{DBLP:journals/tec/GuptaOJZ22}.

The travelling salesperson problem (TSP) is a classical NP-hard computer science problem, where the goal is to generate a tour where the tour length is minimised and all cities are visited exactly once before returning to the starting city. Solving the TSP can be done using both exact and heuristic methods. State-of-the-art exact methods such as Concorde \cite{applegate2006traveling} can solve single instances up to $10000$ nodes in reasonable time with supercomputers, however practically, above a few thousand nodes, solving multiple instances is not viable.
On the other hand, evolutionary state-of-the-art solvers such as EAX \cite{DBLP:journals/informs/NagataK13} and LKH-2 do not find exact solutions but rather approximations in a drastically shorter time frame and are thus viable for these larger problems. While EAX is stronger for large instances, LKH-2 and its extensions are competitive, sometimes beating EAX for problems fewer than around $100000$ nodes. LKH-2 is based on the Lin-Kernighan variable depth search, where at each step the algorithm considers swapping edges of increasing length until no improvement is found while using a 1-tree lower bound to guide the search process. EAX works by maintaining a population of tours and iteratively combining pairs of parent solutions via edge assembly crossover, which creates offspring by selectively swapping segments of edges between parents, preserving high quality building blocks while introducing diversity.
However, running EAX independently from scratch on multiple similar routing problems ignores potentially transferable structural knowledge, wasting computational effort. By integrating multitasking with EAX and creating MT-EAX, we ask whether a) MT-EAX solves problems more efficiently than EAX, and b) MT-EAX yields better solutions than EAX. As MT-EAX solves $k$ problems in one run but for the same compute, for fairness we strictly scale MT-EAX's generation limit and population to evaluate performance under fixed computational budgets. 

This paper shows that MT-EAX in the generation-scaled case is more effective compute-wise than EAX for the same solution quality. This is because the MT-EAX framework allows for parallel solving of instances, injecting increased diversity through similar-instance crossover which outperforms deeper, isolated searching in the early generations. However, when scaling is done by population, the results are more negative, as the larger global population does not improve diversity enough to offset the penalty of fewer generations. The advantage of MT-EAX is significantly extended by utilising a decoupled variation (Decoupled MT-EAX), which acts as a computational cutoff, ensuring that the early-time multitasking advantage is preserved before isolating instances to cleanly reach final convergence.
These results are directly related to the limitations of effective transfer, seen in \cite{Wigney2025Transfer} which shows that even with the most optimal node matching, transferred tours are on average $1.6$ times longer than instance-specific optimal solutions. It also aligns with the results in \cite{DBLP:conf/gecco/Don0025}, confirming that the advantage of multitasking is primarily for limited computational budgets. We also observe that the structural similarity between instances dictates performance, where the geometric distribution of the instances has the largest and most significant effect. 
Overall our results show that MT-EAX can provide better solutions than EAX both at convergence and at fixed budgets, although the primary advantage is that for early fixed-time budgets MT-EAX finds significantly better results than EAX.

The paper is structured as follows:
Section 2 outlines the background of the problem, in terms of the TSP, related work and the EAX algorithm. Section 3 provides a description of MT-EAX and justifications for the design. Section 4 outlines the experimental setup and the evaluation metrics. Section 5 describes the experimental results. Section 6 gives a discussion on these results, relating back to both EAX and MT-EAX.

\section{Background}

Traditionally, solving multiple problems, TSP or otherwise, has required solving each problem separately. However, multitasking has been shown to provide both reduced compute times for similar quality solutions and an increased quality for comparable compute times \cite{DBLP:journals/tec/GuptaOF16,feng2023evolutionary}. 

\subsection{Problem Formulation}
An instance $I$ is given as a set $V=\{v_1, \ldots, v_n\}$ of $n$ cities with distances $d(u,v)$ for any pair of cities $(u,v) \in V \times V$. For a given instance $I=(V, d)$, the goal is to compute a permutation $\pi=(\pi_1, \ldots, \pi_n)$ of the cities of $V$ such that
$$
c(\pi) = d(\pi_n, \pi_1) + \sum_{i=1}^{n-1} d(\pi_i, \pi_{i+1})
$$
is minimal. We consider a set of TSP instances $I=\{I_1, \ldots, I_k\}$ where each instance $I_j=(V_j, d_j)$ has $n$ cities and is defined by a set $V_j=\{v^j_1, \ldots, v^j_n\}$ with distances $d_j(u,v)$ for any pair of cities $(u,v) \in V_j \times V_j$. The goal is to find for any instance $I_j$, $1\leq j \leq k$, a permutation $\pi^j=(\pi^j_1, \ldots, \pi^j_n)$ of the cities of $V_j$ such that
$$
c_j(\pi^j) = d_j(\pi^j_n, \pi^j_1) + \sum_{i=1}^{n-1} d_j(\pi^j_i, \pi^j_{i+1})
$$
is minimal. We consider sets of TSP instances $I$ where each instance $I_s$, $ 1\leq s \leq k$ has the same number of cities $n$.

\subsection{Related Work}
Prior work on multitasking for the TSP has focused on deep learning and reinforcement learning techniques \cite{10754652,DBLP:journals/algorithms/SouzaSOOON24,DBLP:journals/constraints/JoshiCRL22} rather than evolutionary algorithms. These approaches do not explicitly transfer solutions from one instance to another, making it difficult to understand mechanistically how transfer operates. Much of the current evolutionary multitasking literature \cite{DBLP:journals/tcyb/LinLTG21,DBLP:journals/tcyb/LinWMGLC24,SUN2023504,DBLP:journals/isci/HuLSM22} focuses on general combinatorial problems or non-TSP problems. They lack an explicit focus on explainable techniques to compare methods, hindering the ability to understand why certain approaches succeed or fail during transfer. 

More explicit transfer work exists \cite{Wigney2025Transfer}, examining how node ordering can be aligned between instances to minimise distances via matching, allowing permutations generated for one instance to be more applicable to another. Multitasking has also been investigated \cite{DBLP:conf/gecco/Don0025} and shown to be effective in the closely related TTP, where for small time budgets it outperforms classical approaches. This suggests that the benefits of multitasking are restricted to certain stages of the optimisation process for these problems. Work on transferring solutions is not always positive, \cite{XuHao2022EMOW} showed that transferring solutions can cause negative transfer where progress degrades compared to running separately. 

\subsection{The EAX Algorithm}
EAX works by maintaining a population of solutions and iterating over them with selection and crossover operators until termination conditions are met. It is initialised by applying 2-opt local search to random tours, producing population size $P$ locally optimal starting solutions. 

Crossover operates by taking two parent tours, $p_A$ and $p_B$. Let $E_A$ and $E_B$ denote their edge sets. The combined multigraph is then defined as $G_{A,B} = (V, E_A \cup E_B)$. It is decomposed into AB-cycles which are alternative cycles using edges from each edge set in turn. Then an E-set is built by selecting a subset of these AB-cycles depending on the stage the algorithm is in. A temporary solution is formed from $p_A$ by removing the $E_A$ edges in the E-set and replacing them with the corresponding $E_B$ edges, which creates one or potentially more subtours. These are greedily merged, minimising the added edge length to connect into a valid tour. This is done $O$ times, where $O$ is the number of offspring.

There are two stages, I and II, which define the strategy selected for constructing these E-sets. For stage I, a single AB-cycle is selected, which closely resembles $p_A$ and thus maintains population diversity. This functions as a localised search. For stage II, the block2 method is used which applies a tabu search to select AB-cycles that form large adjacent segments, which more drastically changes the offspring and functions as a more global search. 

Selection operates by replacing only $p_A$ with the best offspring rather than $p_B$. The selection preserves diversity via preserving entropy, as offspring are evaluated not on tour length alone, but by the improvement in average tour length per unit loss of population edge entropy in an attempt to prevent premature convergence. Stage I and II both continue until no improvement in the best solution has occurred for $1500/O$ generations. Once the average tour length equals the best, the run ends. 

The edge frequency matrix $F(e)$ tracks the number of individuals in the population that use each edge $e$. It drives diversity maintenance and is utilised by the block2 strategy. Standard EAX uses $F(e)$ to strictly guide crossover based on a single instance's population structure, evaluating population diversity via edge entropy $H$. Offspring selection preserves this diversity by balancing tour length improvement vs diversity loss, evaluating solutions via $\Delta L/\Delta H$. In MT-EAX, calculating $F(e)$ from a pooled subset of the population preserves rare cross-instance edges. This explicitly inflates $H$ with low-frequency, structurally compatible edges, altering the selection criteria to permit exploratory offspring and preventing premature convergence.

\section{MT-EAX}

We now introduce our multi-task version of EAX, called MT-EAX. 
The core principle of MT-EAX is to maximise genetic sharing across instances while strictly preventing instance-specific mechanisms from introducing destructive noise. By maintaining a single, global population pooled across all instances, structural knowledge can flow between distinct routing problems during crossover. However, because edge frequencies are highly geometry-specific, merging the edge frequency matrix $F(e)$ or forcing instances to share convergence stages would influence the algorithm's local search capabilities. Therefore, the baseline MT-EAX isolates the edge frequency matrix, convergence detection, and stage transitions per-instance, while allowing for cross-instance parent selection.

MT-EAX solves multiple problems simultaneously as shown in Algorithm \ref{alg:mteax}. In the baseline configuration, individuals, and thus instances, are selected uniformly at random for crossover, and there is no explicit population transfer. 
Outside of the baseline, several extensions were considered. These include modifying the selection operator, sharing edge frequencies, directing explicit transfer, and utilising decoupled execution.
The shared edge frequency modification differs from Algorithm \ref{alg:mteax} by generating the edge frequencies for the entire global population, rather than strictly from the parents' instance. We evaluated scenarios with warm-up periods to allow the algorithm to build locally optimal structures first.
Sampling from a diverse multi-instance pool explicitly increases early-stage genetic diversity forcing exploration of a broader region of the search space before converging on local structures.

In the directed parent selection modification, selection dictates crossover pairings. We applied Pearson similarity thresholds (calculated between instance distance matrices) at various $r$, as well a same-distribution constraint. Parent B was sampled continuously until the criteria were met or the pool of valid parents was exhausted.
For explicit instance transfer, individuals were directly injected into foreign instances every $M$ generations following a $W$ generation warm-up. This replaced the worst $s$ individuals in a target instance with the best from a compatible instance. To support this, we also tested kopt-supported transfer, where the best tour from Parent A was locally repaired using 2-opt under Parent B's distance evaluator prior to injection, ensuring only competitive repairs were accepted. Prior work on using tours generated from one instance and applying them to another instance \cite{Wigney2025Transfer} provided a framework for improving the ability to generate tours for one instance but apply them to another. This would potentially allow for the effective transfer of knowledge between instances through the tours themselves. 

Finally, we introduce Decoupled MT-EAX. In this configuration, we execute the baseline MT-EAX for $W$ generations before disabling cross-instance crossover entirely. This effectively turns off multi-tasking after the initial phase, forcing the algorithm to operate as $k$ independent EAX runs to isolate the final convergence process. This was motivated by prior work \cite{DBLP:conf/gecco/Don0025} suggesting that the advantages of multitasking most come in the early stages of the optimisation process.

\begin{algorithm}[t]
\small
\caption{MT-EAX: Multi-Task Edge Assembly Crossover}
\label{alg:mteax}
\begin{algorithmic}[1]
\Require $k$ TSP instances $\mathcal{I}_1, \ldots, \mathcal{I}_k$, global population size $P_{\mathrm{MT}}$, offspring count $O$
\Ensure Best tour found for each instance $\mathcal{I}_i$
\State \textbf{Initialisation:} Generate $P_{\mathrm{MT}}$ individuals, assign individual $j$ to instance $j \bmod k$ via round-robin.
\State Improve each individual with 2-opt local search under its assigned instance evaluator.
\Repeat
\State $r(\cdot) \leftarrow$ random permutation of $\{1, \ldots, P_{\mathrm{MT}}\}$ \Comment{Cross-instance mating allowed}
\For{$i = 1$ to $P_{\mathrm{MT}}$}
\State $p_A \leftarrow x_{r(i)}$, $p_B \leftarrow x_{r(i+1)}$
\State Compute edge frequency matrix $F$ from individuals assigned to $\mathcal{I}_{p_A}$
\State $\{c_1, \ldots, c_O\} \leftarrow \textsc{EAX}(p_A, p_B, F, \mathcal{I}_{p_A})$ \Comment{Crossover under $p_A$'s instance}
\State $x_{r(i)} \leftarrow \textsc{SelectBest}(c_1, \ldots, c_O, p_A)$ \Comment{Replace $p_A$ only if improved}
\State $x_{r(i)}.\text{instance} \leftarrow \mathcal{I}_{p_A}$ \Comment{Preserve instance assignment}
\EndFor
\For{$i = 1$ to $k$} \Comment{Per-instance convergence check}
\If{$\textsc{Stagnated}(\mathcal{I}_i, \text{stage}=1)$}
\State Transition instance $\mathcal{I}_i$ to Stage II (block2 EAX)
\EndIf
\If{$\textsc{Stagnated}(\mathcal{I}_i, \text{stage}=2)$ \textbf{or} population collapsed}
\State Mark $\mathcal{I}_i$ as converged
\EndIf
\EndFor
\Until{all $k$ instances converged}
\State \Return best individual for each instance $\mathcal{I}_i$
\end{algorithmic}
\end{algorithm}

\section{Experiments}

To test MT-EAX, we consider a subset of benchmark instances generated using the same methodology as in \cite{Wigney2025Transfer}. These instances are generated on an $A \times A$ grid of the Euclidean plane $[0, A]^2$ for a given number of nodes $n$. We considered node and axis lengths of $100$, $1000$, and $10000$, with location distributions sampled either from a discrete uniform distribution $\sim U(0, A)$ or a normal distribution $\sim N(A/2, A/6)$. We also consider configurations mixing these distributions.

The instances are saved in the TSPLIB95 format \cite{DBLP:journals/informs/Reinelt91} with the distance type EUC\_2D, which rounds distances to the nearest integer, introducing a roughly $2\%$ error in tour lengths. The ordering of the instances is initially lexicographical, with instances being greedily and exactly matched to the first generated instance. For mixed-distribution configurations, the reference instance for matching is the first uniformly distributed instance.

Given the parameters $P$ for population size and $O$ for offspring, solving $k$ problems taking $G$ generations in EAX consumes a computational budget of $B_e = k \cdot P \cdot O \cdot G$. The primary algorithmic overhead resides in the crossover phase. To ensure a fair comparison under fixed computational budgets, we hold the offspring parameter constant and evaluate three distinct scaling methods. Here, $P_{\mathrm{MT}}$ denotes the global MT-EAX population size. \textbf{Generation Scaling:} MT-EAX utilises a global population of size $P_{\mathrm{MT}}=P$ and runs for $k \cdot G$ generations. The total evaluations are equal ($B_m = k \cdot P \cdot O \cdot G$). Because the population $P$ is distributed across instances, each instance operates with $P/k$ individuals. \textbf{Population Scaling:} MT-EAX utilises a global population of size $P_{\mathrm{MT}}=P \cdot k$ and runs for $G$ generations. The total evaluations remain equal ($B_m = k \cdot P \cdot O \cdot G$). \textbf{Balanced Scaling:} MT-EAX utilises a population of size $P_{\mathrm{MT}}=P \cdot \sqrt{k}$ and runs for $G \cdot \sqrt{k}$ generations, achieving an intermediary balance while maintaining the strict budget ($B_m = \sqrt{k} \cdot P \cdot O  \cdot \sqrt{k} \cdot G$).

Scaling the offspring introduces complex internal interactions and obscures the underlying mechanisms, thus it is held constant.
For generation and population scaling, we test 5 sets of instances ($k \in \{2, 4, 5, 9, 10\}$). For balanced scaling we use only two sets ($k \in \{4, 9\}$) to ensure clean integer scaling. For $k \in \{2, 5, 10\}$ we evaluate permutations of $P \in \{100, 200, 500, 1000 \}$ with offspring $O \in \{ 30, 60 \}$. For $k \in \{4, 9\}$ we evaluate permutations of $P \in \{ 100, 200, 300, 400\}$ with $O \in \{30, 60 \}$ and $P \in \{ 100, 300, 900, 1500, 2700, 3000\}$ with $O \in \{30, 60 \}$ respectively.
The shared edge frequency modification had warm-up periods of $W=\{0,20,50,100\}$ generations. The directed parent selection modification had Pearson similarity thresholds (calculated between instance distance matrices) at $r = \{0.2, 0.5, 0.8, 0.9 \}$. For explicit instance transfer, individuals were directly injected into foreign instances every $M= \{10,20,50\}$ generations following a $W = \{0,20,50\}$ generation warm-up. This replaced the worst $s=\{1,5,10\}$ individuals.

We restrict analysis to instances within fixed configurations (identical axis length, node count, and matching type) to isolate factors. For single-distribution setups, the first $k$ instances are utilised. Larger $k$ sets are strict supersets of smaller $k$ sets, ensuring performance is not driven entirely by instance geometry. We execute 30 independent trials per configuration, allowing runs to reach full convergence. Trials utilise deterministic seeds identical to the standard EAX implementation to ensure direct reproducibility.

\subsection{Evaluation Metrics}

We analyse two primary metrics: quality gap and compute saved. These are benchmarked strictly against standard EAX rather than exact Concorde solutions, as this isolates the performance delta introduced by MT-EAX operations, and exact resolution of $10000$-node instances across our exhaustive benchmark set is computationally impractical.

Let $c_{\text{EAX}}$ and $c_{\text{MT}}$ denote the median final tour costs for EAX and MT-EAX respectively across 30 trials on the same instance. The quality gap is defined as:
\begin{equation}
  \Delta Q = \frac{c_{\text{EAX}} - c_{\text{MT}}}{c_{\text{EAX}}} \times 100\%
  \label{eq:quality_gap}
\end{equation}

A positive $\Delta Q$ indicates MT-EAX found a shorter tour, a negative value indicates standard EAX found a shorter tour. Win rate defines the fraction of trials where MT-EAX strictly outperformed EAX. We report median tour costs, which is consistent with the non-parametric Wilcoxon signed-rank test utilised for significance testing.
Let $C$ denote the median generation at which standard EAX reaches final convergence (triggering the natural termination condition of no improvement for $1500/O$ generations). The compute saved at checkpoint $G$ is calculated as:
\begin{equation}
  \text{Compute Saved}(G) = \frac{C - G}{C} \times 100\%
  \label{eq:compute_saved}
\end{equation}
A positive percentage signifies that terminating MT-EAX at generation $G$ consumes less total computational overhead than executing EAX to full convergence. For reference under the primary evaluation parameters ($O=60$), standard EAX converges at a median of $C = 137$ generations for $n=1000$ normally distributed instances, and $C = 117$ for uniformly distributed instances.
For generation scaling efficiency, the EAX-equivalent generation is calculated as $G_{\text{MT}}/k$, and the per-instance population is defined as $P/k$. In the context of population scaling analysis, we map the recovery generation, defined as the first checkpoint where the quality gap recovers to $\geq -0.01\%$. In the tabulated results, a star symbol ($\star$) indicates that MT-EAX at that specific checkpoint has matched or exceeded the final converged quality of standard EAX.

We also define the following secondary metrics to capture smaller dynamics. Peak deficit is the maximum negative $\Delta Q$ observed during a run, occurring at the peak deficit generation. For full convergence efficiency, MT conv is the median generation MT-EAX triggers termination. Sprint gap is the quality gap evaluated at an early target generation (e.g. $G=50$), capturing initial progress prior to full convergence. Peak sprint denotes the maximum positive $\Delta Q$ observed prior to full convergence. For decoupled, the cutoff is the generation threshold where instance populations are strictly isolated, and decoupled conv gen defines the median convergence generation for this isolated phase. Base gap and decoupled gap denote final converged $\Delta Q$ without and with decoupling, respectively. Global $P$ defines the total aggregate population across all evaluated instances.
All comparative quality gaps are evaluated via the Wilcoxon signed-rank test. 

\section{Experimental Results}

\begin{table}[t]
\centering\scriptsize
\setlength{\tabcolsep}{2pt}
\caption{MT-EAX generation scaling results ($n=1000$, $O=60$). Values are quality gaps at matched EAX-equivalent checkpoints; win rates are in parentheses. \(\star\) indicates that the MT-EAX checkpoint solution has matched or exceeded the final converged quality of EAX. Positive gaps (MT-EAX wins) are bolded.}
\label{tab:kgen}
\begin{tabular}{llrrrrr}
\toprule
& & \multicolumn{5}{c}{EAX-equivalent generation $G$} \\
\cmidrule(lr){3-7}
$k$ & Dist. & $G=10$ & $G=20$ & $G=50$ & $G=100$ & $G=200$ \\
\multicolumn{2}{l}{Comp. saved (Normal):} & $93\%$ & $85\%$ & $64\%$ & $27\%$ & $-46\%$ \\
\midrule
$2$ & Normal & $\mathbf{+0.069}$ (58) & $\mathbf{+0.217}$ (70) & $\mathbf{+0.366}^\star$ (83) & $\mathbf{+0.157}^\star$ (78) & $+0.000^\star$ (2) \\
    & Uniform& $-0.051$ (44) & $\mathbf{+0.044}$ (54) & $\mathbf{+0.208}^\star$ (71) & $\mathbf{+0.022}^\star$ (62) & $+0.000^\star$ (1) \\
\midrule
$4$ & Normal & $\mathbf{+0.042}$ (54) & $\mathbf{+0.244}$ (66) & $\mathbf{+0.476}$ (80) & $\mathbf{+0.161}$ (81) & --- \\
    & Uniform& $-0.088$ (41) & $\mathbf{+0.105}$ (56) & $\mathbf{+0.307}$ (70) & $\mathbf{+0.036}$ (67) & --- \\
\midrule
$5$ & Normal & $\mathbf{+0.098}$ (57) & $\mathbf{+0.366}$ (70) & $\mathbf{+0.590}$ (84) & $\mathbf{+0.163}$ (79) & $+0.000$ (0) \\
    & Uniform& $-0.146$ (38) & $\mathbf{+0.043}$ (52) & $\mathbf{+0.271}$ (69) & $\mathbf{+0.022}^\star$ (61) & $+0.000^\star$ (0) \\
\midrule
$9$ & Normal & $\mathbf{+0.402}$ (70) & --- & $\mathbf{+0.892}$ (88) & $\mathbf{+0.141}$ (79) & --- \\
    & Uniform& $-0.035$ (48) & --- & $\mathbf{+0.399}$ (75) & $\mathbf{+0.018}^\star$ (57) & --- \\
\midrule
$10$& Normal & $\mathbf{+0.247}$ (62) & $\mathbf{+0.638}$ (71) & $\mathbf{+0.809}$ (83) & $\mathbf{+0.113}$ (71) & $-0.055$ (0) \\
    & Uniform& $-0.128$ (42) & $\mathbf{+0.086}$ (53) & $\mathbf{+0.345}$ (69) & $\mathbf{+0.004}$ (51) & $-0.043$ (0) \\
\bottomrule
\end{tabular}
\end{table}

\begin{table}[t]
\centering\scriptsize
\setlength{\tabcolsep}{2pt}
\caption{MT-EAX population scaling quality deficits relative to EAX at matched generation checkpoints ($n=1000$, $O=60$). Recovery generation defines the checkpoint where the gap recovers to $\ge -0.01\%$. Deficits are bolded for worst-case isolation.}
\label{tab:kpop}
\begin{tabular}{llrrrr}
\toprule
$k$ & Dist. & Peak deficit $\Delta Q$ (\%) & Peak def. gen & Gen $100$ $\Delta Q$ (\%) & Recovery gen \\
\midrule
$2$ & Normal & $\mathbf{-0.861}$ & 80 & $-0.767$ & 250 \\
    & Uniform& $\mathbf{-1.394}$ & 80 & $-1.116$ & 250 \\
\midrule
$4$ & Normal & $\mathbf{-1.888}$ & 100 & $-1.888$ & 500 \\
    & Uniform& $\mathbf{-2.345}$ & 100 & $-2.345$ & 400 \\
\midrule
$5$ & Normal & $\mathbf{-1.962}$ & 100 & $-1.962$ & 500 \\
    & Uniform& $\mathbf{-2.838}$ & 100 & $-2.838$ & 500 \\
\midrule
$9$ & Normal & $\mathbf{-2.117}$ & 100 & $-2.117$ & 750 \\
    & Uniform& $\mathbf{-3.419}$ & 100 & $-3.419$ & 750 \\
\midrule
$10$& Normal & $\mathbf{-2.429}$ & 100 & $-2.429$ & 1000 \\
    & Uniform& $\mathbf{-3.401}$ & 100 & $-3.401$ & 1000 \\
\bottomrule
\end{tabular}
\end{table}

\begin{table}[t]
\centering\scriptsize
\setlength{\tabcolsep}{2pt}
\caption{MT-EAX full convergence efficiency ($n=1000$, $O=60$). Savings and gaps are relative to full-convergence standard EAX. Positive compute savings are bolded.}
\label{tab:convergence_eff}
\begin{tabular}{llrrrrrr}
\toprule
$k$ & Dist. & EAX $C$ & MT conv & EAX-equiv & Saving (\%) & MT final tour cost & Final $\Delta Q$ (\%) \\
\midrule
$2$ & Normal & 137 & 243 & 122 & $\mathbf{+11.3}$ & 17793 & $+2.381$ \\
    & Uniform& 117 & 215 & 108 & $\mathbf{+8.1}$ & 22305 & $+3.013$ \\
\midrule
$4$ & Normal & 137 & 460 & 115 & $\mathbf{+16.1}$ & 18230 & $-0.016$ \\
    & Uniform& 117 & 384 &  96 & $\mathbf{+17.8}$ & 22998 & $+0.000$ \\
\midrule
$5$ & Normal & 137 & 484 &  97 & $\mathbf{+29.3}$ & 18357 & $-0.713$ \\
    & Uniform& 117 & 481 &  96 & $\mathbf{+17.8}$ & 22976 & $+0.096$ \\
\midrule
$9$ & Normal & 137 & 740 &  82 & $\mathbf{+40.0}$ & 18272 & $-0.247$ \\
    & Uniform& 117 & 700 &  78 & $\mathbf{+33.5}$ & 22998 & $+0.000$ \\
\midrule
$10$& Normal & 137 & 801 &  80 & $\mathbf{+41.5}$ & 18244 & $-0.093$ \\
    & Uniform& 117 & 786 &  79 & $\mathbf{+32.8}$ & 22998 & $+0.000$ \\
\bottomrule
\end{tabular}
\end{table}
\begin{table}[t]
\centering\scriptsize
\setlength{\tabcolsep}{2pt}
\caption{Decoupled MT-EAX performance ($n=1000$, $O=60$). Base and Dec. $\Delta Q$ denote final-convergence gaps. Max $\Delta Q$ denotes the maximum positive gap at matched EAX-equivalent checkpoints, with the corresponding generation $G$ in parentheses. Positive gaps and savings are bolded.}
\label{tab:decoupled_mteax}
\begin{tabular}{llrrrrrr}
\toprule
$k$ & Dist. & Cutoff & Base $\Delta Q$ (\%) & Dec. $\Delta Q$ (\%) & Max $\Delta Q$ (\%) & Dec. conv gen & Compute saved (\%) \\
\midrule
$2$ & Normal & 100 & $\mathbf{+2.381}$ & $\mathbf{+2.381}$ & $\mathbf{+2.896}$ (75) & 178 & $\mathbf{+35.0}$ \\
$2$ & Normal & 200 & $\mathbf{+2.381}$ & $\mathbf{+2.381}$ & $\mathbf{+2.560}$ (50) & 224 & $\mathbf{+18.2}$ \\
$2$ & Uniform& 100 & $\mathbf{+3.013}$ & $\mathbf{+3.013}$ & $\mathbf{+3.509}$ (75) & 162 & $\mathbf{+30.8}$ \\
$2$ & Uniform& 200 & $\mathbf{+3.013}$ & $\mathbf{+3.013}$ & $\mathbf{+3.233}$ (75) & 209 & $\mathbf{+10.7}$ \\
\midrule
$4$ & Normal & 100 & $-0.016$ & $-0.016$ & $\mathbf{+1.512}$ (50) & 206 & $\mathbf{+62.4}$ \\
$4$ & Normal & 200 & $-0.016$ & $-0.016$ & $\mathbf{+0.751}$ (75) & 276 & $\mathbf{+49.6}$ \\
$4$ & Uniform& 100 & $+0.000$ & $\mathbf{+1.454}$ & $\mathbf{+2.844}$ (50) & 180 & $\mathbf{+61.5}$ \\
$4$ & Uniform& 200 & $+0.000$ & $\mathbf{+1.433}$ & $\mathbf{+1.968}$ (75) & 250 & $\mathbf{+46.6}$ \\
\midrule
$5$ & Normal & 100 & $-0.713$ & $-0.713$ & $\mathbf{+1.280}$ (30) & 198 & $\mathbf{+71.0}$ \\
$5$ & Normal & 200 & $-0.713$ & $-0.713$ & $\mathbf{+0.821}$ (50) & 268 & $\mathbf{+60.9}$ \\
$5$ & Uniform& 100 & $\mathbf{+0.096}$ & $\mathbf{+0.096}$ & $\mathbf{+1.963}$ (40) & 187 & $\mathbf{+68.0}$ \\
$5$ & Uniform& 200 & $\mathbf{+0.096}$ & $\mathbf{+0.096}$ & $\mathbf{+1.260}$ (50) & 263 & $\mathbf{+55.0}$ \\
\midrule
$9$ & Normal & 100 & $-0.247$ & $-0.247$ & $\mathbf{+1.307}$ (50) & 215 & $\mathbf{+82.6}$ \\
$9$ & Normal & 200 & $-0.247$ & $-0.247$ & $\mathbf{+1.307}$ (50) & 295 & $\mathbf{+76.1}$ \\
$9$ & Uniform& 100 & $+0.000$ & $+0.000$ & $\mathbf{+1.410}$ (50) & 194 & $\mathbf{+81.6}$ \\
$9$ & Uniform& 200 & $+0.000$ & $+0.000$ & $\mathbf{+1.410}$ (50) & 276 & $\mathbf{+73.8}$ \\
\midrule
$10$& Normal & 100 & $-0.093$ & $-0.060$ & $\mathbf{+2.650}$ (20) & 209 & $\mathbf{+84.7}$ \\
$10$& Normal & 200 & $-0.093$ & $-0.069$ & $\mathbf{+2.274}$ (30) & 288 & $\mathbf{+79.0}$ \\
$10$& Uniform& 100 & $+0.000$ & $+0.000$ & $\mathbf{+2.863}$ (20) & 197 & $\mathbf{+83.2}$ \\
$10$& Uniform& 200 & $+0.000$ & $+0.000$ & $\mathbf{+2.369}$ (30) & 279 & $\mathbf{+76.2}$ \\
\bottomrule
\end{tabular}
\end{table}

\begin{table}[t]
\centering\scriptsize
\setlength{\tabcolsep}{2pt}
\caption{Population starvation gradient quality gaps ($k=10$, Normal, $n=1000$). Sprint gaps compare matched EAX equivalent checkpoints while convergence gaps are relative to full-convergence standard EAX. Positive sprint gaps are bolded.}
\label{tab:starvation_gradient}
\begin{tabular}{rrrrrr}
\toprule
& & \multicolumn{2}{c}{Sprint gap (\%, Gen 50)} & \multicolumn{2}{c}{Conv gap (\%, Final)} \\
\cmidrule(lr){3-4} \cmidrule(lr){5-6}
Global $P$ & Per-inst $P/k$ & $O=30$ & $O=60$ & $O=30$ & $O=60$ \\
\midrule
 100 &  10.0 & $\mathbf{+0.427}$ & $\mathbf{+0.793}$ & $-0.324$ & $-0.351$ \\
 200 &  20.0 & $\mathbf{+0.523}$ & $\mathbf{+0.972}$ & $-0.088$ & $-0.071$ \\
 500 &  50.0 & $\mathbf{+0.591}$ & $\mathbf{+0.992}$ & $-0.016$ & $-0.016$ \\
1000 & 100.0 & $\mathbf{+0.589}$ & $\mathbf{+0.936}$ & $-0.005$ & $-0.011$ \\
\bottomrule
\end{tabular}
\end{table}
For the $n=100$ node case, there is no statistical difference between any result, as EAX converges in a median of $3$ generations on these scale instances, leaving no window for parallel advantages. For the $n=10000$ cases, we got similar results to the $n=1000$, with larger generational lengths and thus compute limitations for the exhaustive test cases. Thus, our analysis focuses strictly on the $1000$-node benchmark sets. 

The baseline MT-EAX results significantly outperformed all structural extensions. The primary performance drivers were identified strictly as the scaling method, node distribution, and instance count. Consequently, for our primary results, we evaluate the baseline configuration utilising an offspring size of $O=60$, which we later demonstrate provides a buffer against population starvation.

\subsection{Scaled Generations}

Table \ref{tab:kgen} details the performance of generation-scaled MT-EAX over computational time. Following the initial $10$ generations, the win rate becomes substantially positive, yielding compute savings. While compute saved peaks proportionally at generation $20$, the magnitude of the win rate and quality gap peak near generation $50$ for virtually all parameters. The advantage is fully eroded by generation $200$, at which point the median gap returns to essentially $0\%$.

Table \ref{tab:kgen} shows a distinct geometric bias where clustered, normally distributed instances exploit MT-EAX crossover better than uniformly distributed instances, maintaining an approximately $15\%$ higher win rate throughout the early search phase.  Uniformly distributed instances remain highly consistent, indicating that normal distribution structures are more sensitive to cross-instance crossover parameters.

The compute efficiency displayed in Table \ref{tab:convergence_eff} highlights the strength of the generation-scaling mechanism. At EAX-equivalent generation $50$ (a $64\%$ compute save versus full EAX convergence at $C=137$), MT-EAX at $k=9$ (Normal) is already $+0.892\%$ better than standard EAX at the equivalent checkpoint. More importantly, halting the algorithm at EAX-equivalent generation $100$ (a $27\%$ compute save) with $k=9$ yields a $+0.141\%$ improvement over EAX at the equivalent checkpoint. 

\subsection{Scaled Population and Balanced}

\begin{figure}[t] 
    \centering
    \includegraphics[width=\textwidth]{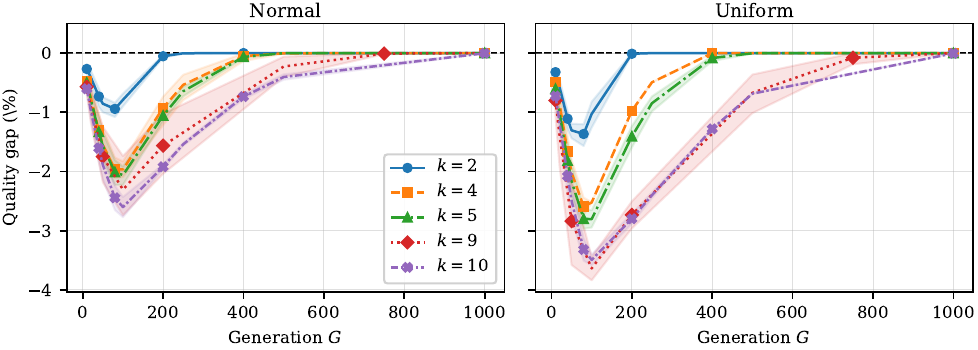} 
    \caption{Quality deficit of population-scaled MT-EAX in early generations ($n=1000$).}
    \label{fig:qual}
\end{figure}

Conversely, when scaling the global population size, Table \ref{tab:kpop} and Figure \ref{fig:qual} demonstrate negative results in the early stages of optimisation. No compute is saved and the deficit is worst near generation $100$, in contrast to the generation-scaled experiments which peaked in this exact window. There is a negative correlation between the number of instances $k$ and the depth of the quality deficit. Furthermore, uniformly distributed instances have worse early penalties than normally distributed instances. The balanced scaling followed identical trends, showing that partial generation scaling fails to overcome the penalty of partial population scaling.

Figure \ref{fig:qual} shows that while the recovery trajectory is consistent, the depth of the early deficit scales with geometry and instance count. For normally distributed instances at $k=9$, the largest gap is $-2.117\%$, whereas uniformly distributed instances drop to $-3.419\%$. The mechanism driving this deficit is cross-instance parent incompatibility. Because normally distributed instances possess localised geometric similarities, cross-instance parents remain more compatible. The deficit scales proportionally with $(k-1)/k$, representing that crossovers forced to utilise an incompatible second parent. For $k=2$, $50\%$ of parents are incompatible, yielding a gap of $-0.861\%$. For $k=9$, $89\%$ are incompatible, moving the gap to $-2.117\%$. Notably, recovery is complete by final convergence. All configurations eventually reach parity ($0.00\%$), demonstrating that $k \times \text{pop}$ scaling ultimately resolves into $k$ independent EAX runs. 

\subsection{Decoupled MT-EAX}

To address the compute inefficiency observed in the late stages of the generation-scaled baseline (Table \ref{tab:convergence_eff}), we evaluated Decoupled MT-EAX. Table \ref{tab:decoupled_mteax} shows that decoupling instances at fixed generation cutoffs had minimal impact on the final global quality gap for most configurations, as the gaps remain virtually identical outside of $k=4$ Uniform. However, the convergence speedups achieved by Decoupled MT-EAX demonstrated a significant computational advantage. In the baseline MT-EAX, the algorithm requires more generations to converge when using incompatible cross-instance edges. Furthermore, Table \ref{tab:decoupled_mteax} demonstrates that for instance sets ($k \in \{2, 4, 5\}$), Decoupled MT-EAX usually outperforms standard EAX at final convergence, yielding a positive final quality gap (up to $+3.013\%$ for $k=2$ Uniform). It maintains a positive peak early-stage advantage across all configurations. It is only in some configurations that a negligible deficit emerges. By severing cross-instance interaction, the algorithm rapidly detects convergence. For example, for $k=10$ (Normal, Cutoff $100$), Decoupled MT-EAX reduces the convergence time from $801$ generations in the baseline down to $209$ generations. This increases the compute saved from $+41.5\%$ to $+84.7\%$, while preserving a peak early-stage advantage of $+2.650\%$ and yielding a very similar final tour cost.

\section{Discussion}

The empirical results demonstrate a clear performance gap based on the available computational budget. In fixed-budget environments, generationally scaled MT-EAX is significantly more effective compute-wise than standard EAX. The mechanism driving this improvement is the injection of structural diversity via similar-instance crossover, which outperforms local search in the early generations. Conversely, when scaling is applied to the population size, the results are strongly negative, a large global population does not offset the penalty of cross-instance incompatibility. 

Under generation scaling, MT-EAX maintains dominance up to an intersection point near generation $200$, where parity with standard EAX is achieved. The geometry, through density, controls the performance in this window: clustered, normally distributed instances exploit the shared population more effectively than uniform distributions, indicating that structural similarities improve crossover viability. MT-EAX operates with a restricted population per instance ($P/k$) but cycles through genetic material $k$ times faster than standard EAX for an equivalent compute budget. This turnover process allows the population to quickly escape poor configurations, which gives the early compute savings. However, as optimisation progresses, the lack of per-instance population depth restricts the localised search required to finalise incremental improvements. Standard EAX is slower initially, but its large, isolated population maintains the diversity necessary to construct final optimal tours.

\begin{figure}[t] 
    \centering
    \includegraphics[width=0.8\textwidth]{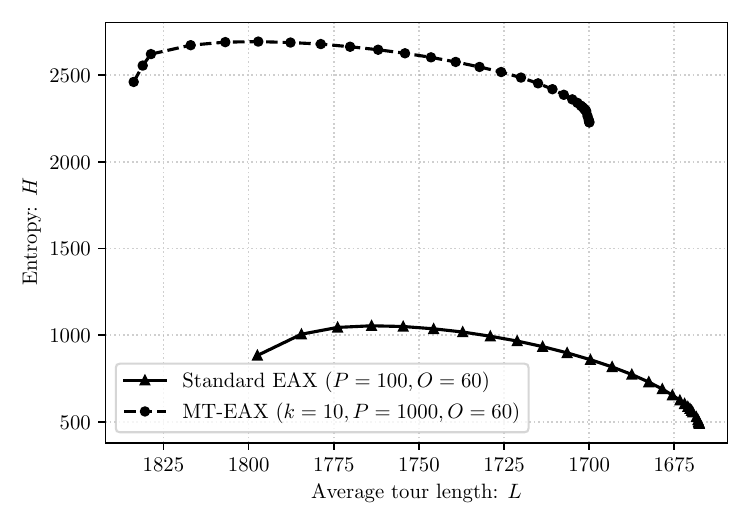}
    \caption{Trajectory of population edge entropy $H$ against average tour length $L$. MT-EAX maintains significantly higher structural diversity at equivalent tour lengths in early generations.}
    \label{fig:entropy_trajectory}
\end{figure}

Table \ref{tab:starvation_gradient} quantifies this population starvation dynamic. As the per-instance population $P/k$ scales from $10.0$ to $100.0$, the final convergence gap moves from $-0.351\%$ to $-0.011\%$. This demonstrates a directly proportional relationship between instance population depth and final convergence capability. With ($P/k=10$), an offspring size of $O=60$ provides twice as many iterations to locate a viable crossover product, nearly doubling the early-stage advantage ($+0.793\%$) compared to $O=30$ ($+0.427\%$).
Large $k$ final quality gaps in the Decoupled MT-EAX validate the expectation that cross-instance noise and a lack of genetic diversity due to the starved per-instance population cause this deficit at convergence. However, in fixed-budget contexts, this late-stage deficit represents a trade-off. For state-of-the-art solvers processing $1000$-node TSP instances, quality gaps are sub-$1\%$. MT-EAX sacrifices a negligible fraction of late-stage optimality to gain large early-stage improvements in quality (up to $+3.509\%$), while also reducing total compute by up to $84.7\%$.

In the early stages ($0$ to $100$), edge frequencies operate as noise, hence blending them yields no optimisation benefits. Explicit transfer, with and without kopt, failed because redistributing converged individuals between disparate geometric spaces degrades the EAX crossover logic. Although optimal node matching yields a $1.635$ length ratio against converged solutions, EAX populations  surpass this within $20$ generations. Transferring an optimally matched solution is unable to be competitive.

The early-stage computational advantage is linked to the preservation of edge entropy $H$ as seen in Figure \ref{fig:entropy_trajectory}. For EAX, $F(e)$ rapidly focuses on local structures, minimising $H$ and halting exploration. Cross-instance crossover introduces compatible structural diversity, artificially inflating $H$ with low-frequency edges. By pooling populations, MT-EAX forces $H$ to remain higher than isolated standard EAX at equivalent average tour lengths $L$. This alters the $\Delta L/\Delta H$ selection criteria, allowing the algorithm to accept exploratory offspring and accelerating early-stage cost reductions. While MT-EAX accelerates early-stage cost reductions, it can converge prematurely at a higher final cost compared to standard EAX. The restricted per-instance population lacks the genetic depth required for the final improvements, validating the Decoupled MT-EAX configuration that can reach or exceed the converged tour lengths from EAX.

\section{Conclusions}
Parallel MT-EAX dominates isolated EAX execution during the early stages of optimisation, making it highly effective for strict, fixed-budget computational environments. Under fair generation scaling, MT-EAX gives significant early-stage quality improvements while simultaneously saving $60\%$ to $90\%$ of necessary compute. 
This computational advantage is further extended by utilising Decoupled MT-EAX. By stopping cross-instance interaction after an initial phase, the algorithm preserves the substantial early-stage speedups (saving up to $84.7\%$ over standard EAX) while allowing individual instances to specialise, often matching or strictly outperforming EAX's final global convergence quality. 
We demonstrated that direct solution transfer is fundamentally restricted, as the native EAX populations rapidly beat the relevance of geometrically transferred solutions within narrow generational windows. The strongest determinant of MT-EAX success is instance geometry, with clustered, normally distributed geometries exploiting cross-instance structures up to $15\%$ more effectively than uniform distributions. Population starvation dictates the bounds of this advantage, governing a trade-off between rapid early optimisation and final global convergence. Future work includes investigating mechanisms to pre-solve reference instances to seed populations for the earliest stages of the multi-task sequence.

\bibliographystyle{unsrt}
\bibliography{final-ref}

\end{document}